\documentclass[11pt]{article}
\usepackage{booktabs}
\usepackage{tabularx}
\usepackage{array}
\usepackage{makecell}
\usepackage[utf8]{inputenc}
\usepackage[T1]{fontenc}
\usepackage{lmodern}
\usepackage{geometry}
\usepackage{amsmath,amssymb}
\usepackage{graphicx}
\usepackage{subcaption}
\usepackage{booktabs}
\usepackage{array}
\usepackage{longtable}
\usepackage{lineno}
\usepackage[round]{natbib}
\usepackage{xcolor}
\usepackage{hyperref}
\usepackage{wrapfig}
\usepackage{placeins} 
\graphicspath{{figures/}}
\usepackage{float}
\nolinenumbers

\title{Pointwise is Pointless?\\
A Multimodal Ablation Study for Precipitation Nowcasting with Graph Neural Networks}

\author{Oph{\'e}lia Miralles$^{1}$ \and M{\'a}t{\'e} Mile$^{1}$\\
\and Christoffer Artturi Elo$^{1}$\\
\and Thomas Nipen$^{1}$ \\
\and Ivar Seierstad$^{1}$\\
[0.5em]
$^{1}$Norwegian Meteorological Institute, Oslo, Norway}
\date{}

\begin{document}
\maketitle

\begin{abstract}
Precipitation nowcasting is usually evaluated as a spatial field problem, but many observations that can improve short-range forecasts are point measurements. This creates a practical tension: surface stations provide direct information on precipitation reaching the ground, whereas radar provides the dense spatial target needed for warnings and downstream applications. In this article, we use a multimodal graph neural network nowcasting system over the Nordic radar domain to study this tension empirically. The model predicts rain rate every five minutes up to two hours ahead and is trained with combinations of radar history, MEPS numerical weather prediction, Netatmo surface observations, MSG satellite channels, stochastic noise, and CRPS-based ensemble losses. We compare radar-only, NWP-informed, station-informed, satellite-informed, noise-augmented, and CRPS-based configurations using radar-grid, station, onset, oracle, displacement, and amplitude diagnostics. The ablation shows that different data sources improve different forecast properties. MEPS stabilises the radar-only forecast, Netatmo observations improve local station-based and onset diagnostics, and satellite predictors reduce some station-level biases but can also favour premature rain activation when used deterministically. CRPS-based configurations give the most consistent radar-grid improvements, while the combined satellite and CRPS configuration gives the lowest overall oracle/DAS score. The results therefore do not indicate that point observations are uninformative. Instead, they show that improvements at the local station and dense radar-grid skill are distinct verification targets. This supports a view of precipitation nowcasting in which sparse observations provide valuable local constraints, but their impact on spatially coherent radar-like fields depends on the training objective and the representation of observation support.
\end{abstract}

\noindent\textbf{Significance statement.} 
The sentence ``pointwise is pointless'' is intentionally provocative. Point observations are the opposite of pointless: they are often the most direct measurements of surface precipitation. The problem is that pointwise losses alone do not explain how a local wet or dry signal should modify a neighbouring precipitation field. This study shows, using a large Nordic nowcasting ablation, that simply adding stations to a deterministic graph neural network can improve the model at the station locations without consistently improving the spatial rain-rate map. The results motivate a shift from direct multimodal regression toward generative assimilation, in which sparse observations constrain an ensemble of radar-like precipitation fields. A follow-up paper will investigate generative assimilation from sparse observations to dense high-resolution precipitation fields.

\section{Introduction}
Nowcasting is the prediction of weather with local detail over the next few minutes to a few hours. For precipitation, this range is particularly important for warnings, hydrology, transport, aviation, and outdoor activities because the damaging part of rainfall often comes from small, rapidly evolving structures that are generally difficult to resolve at the beginning of operational numerical weather prediction (NWP) integrations. Radar extrapolation methods therefore remain strong operational baselines at the shortest lead times because they use the most recent spatial observation of precipitation directly. Their limitations are also well known: they tend to advect what already exists and therefore can struggle with growth, decay, or initiation of convective precipitation \citep{predict1994, predict2006, haiden2011integrated, ipol201326, sideris2014, 2020siderisnowprecip}.

Deep-learning (DL) nowcasting methods aim to learn these nonlinear changes from data. Recent work has shown that neural networks can produce skillful radar nowcasts \citep{agrawal2019machine, sonderby2020metnet, Franch_et_al_2020, rs14163890, andrychowicz2023deep, zhang2023skilful, miralles2025deeplearning}. Probabilistic and generative DL forecasts can represent multiple plausible future evolutions and tend to alleviate the excessive spatial smoothing often produced by deterministic regression models \citep{ravuri2021skilful, leinonen_latent_2023, franchgpt2025}. Graph neural networks (GNNs) are particularly attractive in operational weather forecasting because they can also be used with irregular meshes and heterogeneous data sources, and are now widely used in data-driven weather prediction \citep{lam2022graphcast, price2024gencast, lang2024aifs, lang2024aifscrps, nipen2024regional, miralles2025deeplearning}. Multimodal GNNs are therefore well suited to nowcasting problems in which radar grids, satellite imagery, NWP fields, topography, and surface stations must be used together. By mapping these heterogeneous sources into a common latent representation, they provide an alternative to the lengthy, partially hand-designed preprocessing chains often required to reconcile differences in resolution, spatial support, topology, and update frequency in traditional nowcasting systems and recent deep-learning nowcasting models.

The use of surface stations raises a specific difficulty. Stations provide direct measurements of precipitation reaching the ground, whereas radar provides spatially coherent but indirect estimates aloft.  In the Nordic
region, surface observations are also abundant, and crowd-sourced or private station networks can
provide tens of thousands of additional reports \citep{nipen2020adopting}. In complex terrain, radar beams can overshoot low-level precipitation, observe hydrometeors that evaporate before reaching the surface, or miss shallow precipitation forming below the beam. Stations can therefore contain information that is absent from the radar field, but only at sparse and highly localised locations. This creates a practical question for neural nowcasting: when precipitation is observed by both radar and surface stations, how should a model combine these sources, especially when they disagree?

This question cannot be answered by a single score. Radar-derived rain rate is noisy and uncertain, especially in complex terrain, but it provides the spatial object that the nowcaster is expected to predict. A station observation can indicate that precipitation is reaching the ground at one location, but many spatially coherent radar-like fields remain compatible with that single point measurement. The difficulty is therefore not only architectural, but also concerns the definition of the learning target and the verification support. In this study, we deliberately remain within a direct supervised nowcasting framework and ask how far such a framework can be pushed before this ambiguity becomes limiting.

We use the term ablation in the machine-learning sense: a controlled sequence of experiments in which data sources or training components are added or removed while keeping the remaining model configuration as similar as possible. This differs from classical data-denial experiments in data assimilation, where observations are withheld from an assimilation cycle and their impact is mediated through explicitly specified error covariances. Here, ablation is used to diagnose how each source or objective changes the learned forecast. In particular, it allows us to test whether point observations improve local station verification, whether these gains propagate to the surrounding radar grid, and whether different objectives change the balance between rain occurrence, intensity, and spatial structure. The ablation therefore provides an empirical diagnostic of the support mismatch between point observations and dense radar targets, rather than a complete solution to the observation-fusion problem.

We provide a controlled ablation of multimodal observation fusion for radar precipitation nowcasting over the Nordics. The model combines dense radar targets, NWP predictors, satellite channels, surface stations, stochastic perturbations, and spatially aware losses to obtain forecasts that are both spatially coherent and locally accurate. The specific goals are: i) to establish a graph-based radar nowcasting baseline that is competitive with an optical-flow benchmark up to two hours ahead; ii) to quantify the marginal effect of adding NWP, station observations, satellite channels, and stochastic noise; iii) to diagnose whether remaining errors are dominated by displacement, timing, or intensity; and iv) to identify whether point observations improve spatial fields or mainly the exact locations at which they are observed. The results are intended to motivate a follow-up study in which point observations are not treated as additional deterministic targets, but as constraints in a guided assimilation framework inspired by \citep{NEURIPS2024_9f94298b} to generate radar-like precipitation initial states before feeding them to the two-hour nowcasting model.

This study is structured as follows. Section~\ref{sec:sec2} describes the data sources, the nowcasting task, the graph neural network, the sparse multimodal loading, loss functions, and ablation design. Section~\ref{sec:verification} defines the verification framework used to separate radar-grid skill, station agreement, rain-onset behaviour, displacement errors, and amplitude errors. Section~\ref{sec:results} presents the ablation results, and Section~\ref{sec:discussion} discusses their implications for future work with generative assimilation of sparse observations.

\section{Data and experimental setup}\label{sec:sec2}
\subsection{Data sources}
\label{sec:data}
\begin{table}[ht!]
\centering
\caption{Data sources used in the ablation study. The training period is May 2020--August 2024, the remaining months of 2024 are used for validation, and 2025 is used for testing.}
\label{tab:data_sources}
\footnotesize
\begin{tabularx}{\textwidth}{lclX}
\toprule
Source & Resolution/frequency & Period used & Role in the experiments \\
\midrule
Nordic radar composite & $\sim$1km, 5min & 2020--2025 & Main dense precipitation target and radar-history input. \\
MEPS & $\sim$2.5km, 6h & 2020--2025 & Regional NWP predictors providing dynamical and thermodynamic context. \\
Netatmo & Point observations, 5min & 2020--2025 & Sparse surface observations used as inputs and station targets. \\
MSG/SEVIRI & Satellite grid, 15min & 2020--2025 & Cloud and radiance information used in satellite-informed runs. \\
\bottomrule
\end{tabularx}
\end{table}
\newpage
\subsubsection{Nordic radar composite}
\begin{wrapfigure}{l}{0.45\columnwidth}
  \centering
  \includegraphics[width=0.45\columnwidth]{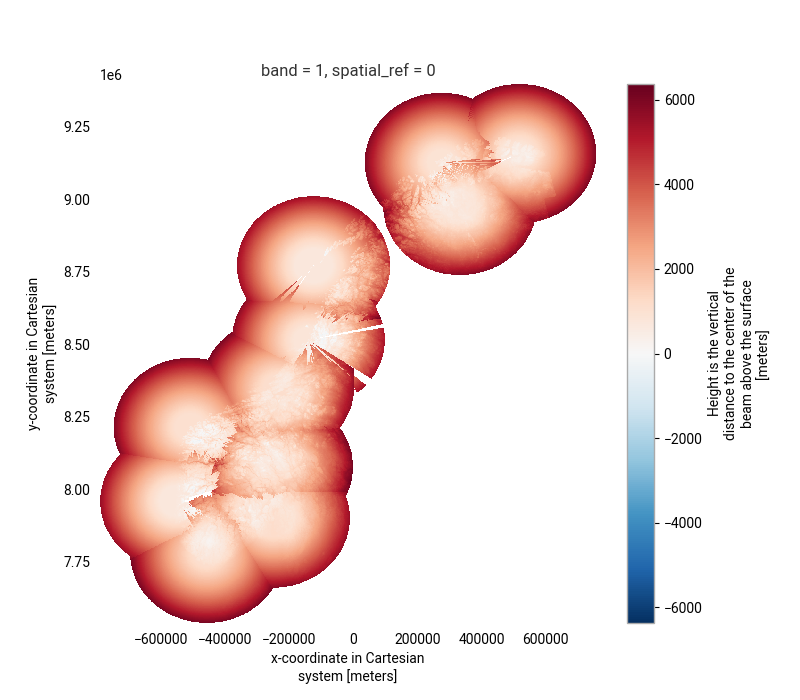}
  \caption{Radar sampling height over the Nordic domain.}
  \label{fig:radar_height}
\end{wrapfigure}

The primary target is a Nordic radar composite covering Norway, Sweden, and Finland at approximately 1~km resolution and 5~min frequency. The nordic composite is preferred over the OPERA composite covering Europe \citep{opera} for the present experiments because it uses regional quality-control procedures and provides better solid precipitation estimates over the Nordic domain. Radar coverage is highly non-uniform. Radars are far apart, the beam is frequently high above the surface because of terrain blocking (Figure~\ref{fig:radar_height}), and precipitation can form below the beam or evaporate before reaching the ground. These limitations make the radar target imperfect, but it is still the only observation source in the current system that provides a dense precipitation field at the spatial and temporal scale required by the nowcaster.

\begin{wrapfigure}{r}{0.45\columnwidth}
  \centering
  \includegraphics[width=0.45\columnwidth]{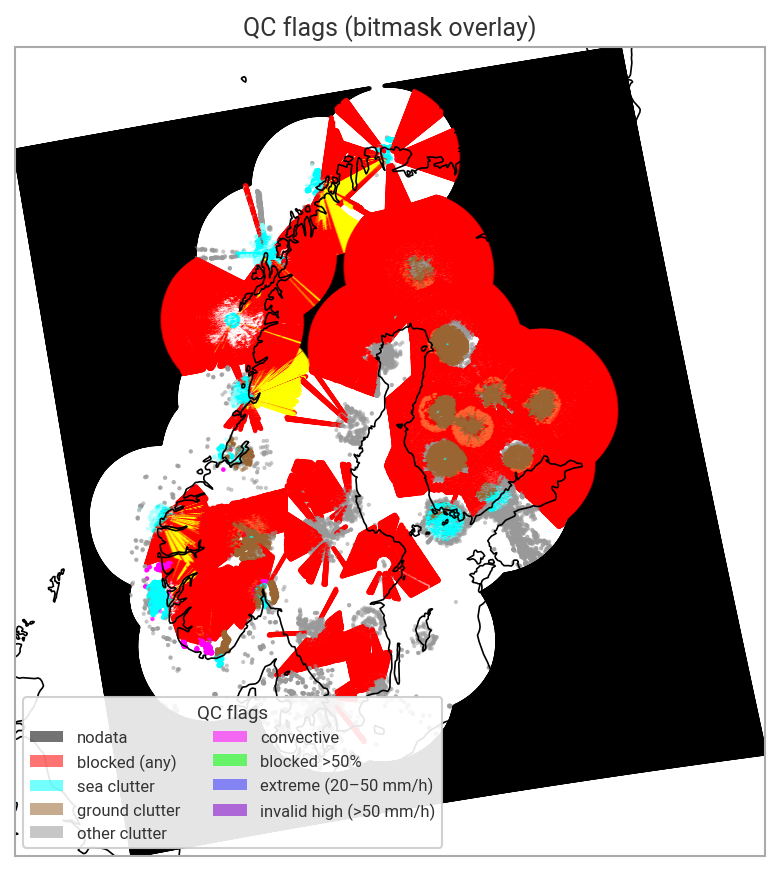}
  \caption{Example of the Nordic radar quality-control bitmask on 2 October 2020. Several binary radar-quality fields, including no-data areas, beam blocking, clutter, convective classification, extreme rates, and invalid high values, are compressed into a single integer flag field. The encoded flags are used both to mask unreliable radar targets and to retain information on the trustworthiness of each pixel.}
  \label{fig:qc_flags}
\end{wrapfigure}

The radar products include several quality-control variables that describe missing coverage, beam blocking, clutter, convective classification, and physically suspicious precipitation rates (Figure~\ref{fig:qc_flags}). Storing each of these fields as a separate dense Boolean array would be inefficient, especially for multi-year archives on a high-resolution, five-minute radar grid ($\simeq$4M grid points). We therefore compress the radar quality information into a single integer bitmask. Each bit of this mask corresponds to one quality-control condition. For the Nordic radar composite, the first bits encode the native radar flags, including no-data, beam blocking, low- and high-elevation sampling, sea clutter, ground clutter, other clutter, and convective classification. 

Additional bits are then derived from the rain-rate field and blocking percentage, marking pixels with more than 50\% beam blocking, extreme but retained rain rates between 20 and 50mm h$^{-1}$, and invalid rain rates above 50mm h$^{-1}$. Pixels affected by hard quality-control failures, such as no-data, clutter, severe blocking, or invalid high rain rates, are set to missing in the rain-rate field, while the corresponding information is retained in the bitmask. This representation is used primarily for compression and storage, but it also preserves information that may be useful during training: the decoded flags can be used to down-weight uncertain radar pixels in the loss, mask unreliable targets, or be passed to the neural network as embedded quality-control inputs. For reproducibility, Appendix~\ref{app:qc_flags} gives pseudo-code for the construction and decoding of the compressed radar quality-control bitmask used throughout the experiments.

\newpage
\subsubsection{Numerical weather prediction}
The MetCoOp Ensemble Prediction System (MEPS) introduced in \citet{muller2017, frogner2019a} is used as the regional NWP source. The NWP input is the latest available MEPS state on the operational regional grid. We do not interpolate MEPS fields to the 5min radar frequency, as doing so would substantially increase data loading during training without adding independent NWP information at that timescale; this choice is discussed further in Section~\ref{sec:sparse}. The fields provided to the model are 2-meter temperature, 6-hour accumulated total precipitation, low and medium cloud cover, land--sea mask, mean sea-level pressure, surface geopotential, and the 10-meter wind components. These variables provide information on the recent thermodynamic state, accumulated precipitation, cloud structure, surface forcing, and near-surface flow available from the regional NWP system at initialization time. 

The role of NWP in the present ablation is to test whether a deep learning nowcaster can combine the short-term spatial precision of radar with the dynamical information contained in the regional model. In the two-hour range, this is especially relevant for precipitation growth and decay, advection across radar gaps, and large-scale frontal precipitation.

\subsubsection{Station observations}
In-situ observations are obtained from the dense private station network called Netatmo \citep{netatmo2026weatherdata}. Such observations have already been shown to add substantial value to operational post-processed forecasts disseminated through Yr, the public weather forecasting service operated by the Norwegian Meteorological Institute \citep{nipen2020adopting,yr2026}. The variables considered include deaccumulated 10-min rain rate, temperature, humidity, pressure, wind components, and the number of neighbouring stations, used as a lightweight proxy for the quality-control information described by \citet{nipen2020adopting}. The precipitation observations are sparse in space, heterogeneous in quality, and not always available at the same temporal frequency as the radar composite. They are therefore represented as sparse graph nodes rather than interpolated to the full radar grid before ingestion.

\subsubsection{Satellite and topography}
Geostationary satellite channels from Meteosat Second Generation (MSG) Spinning Enhanced Visible and Infrared Imager (SEVIRI) are used in some experiments. The channels include infrared, water vapour and visible bands that contain information about cloud structure, cloud-top temperature, low clouds, and surface properties under clear skies. Satellite information is available at lower temporal and spatial resolution than the radar composite, but it can provide predictors for growth and decay that are not visible in recent radar history alone.

Topographic information is included through the MEPS surface geopotential field, used here as a static predictor on the approximately 2.5~km MEPS grid. This provides the model with large-scale information on terrain height and coastal geometry, which is particularly relevant over Norway, where fjords, mountains, coastlines, and radar beam geometry strongly affect both precipitation occurrence and radar quality. We do not include an additional high-resolution subgrid topography encoder in the final ablation. A 500 meter resolution elevation encoder was introduced during model development, but it increased the complexity and computational cost of the multimodal graph without producing a commensurate improvement in the verification scores considered.

\subsection{Graph Neural Network Architecture}
\label{sec:arch}
\begin{figure}[t]
\centering
\includegraphics[width=0.8\textwidth]{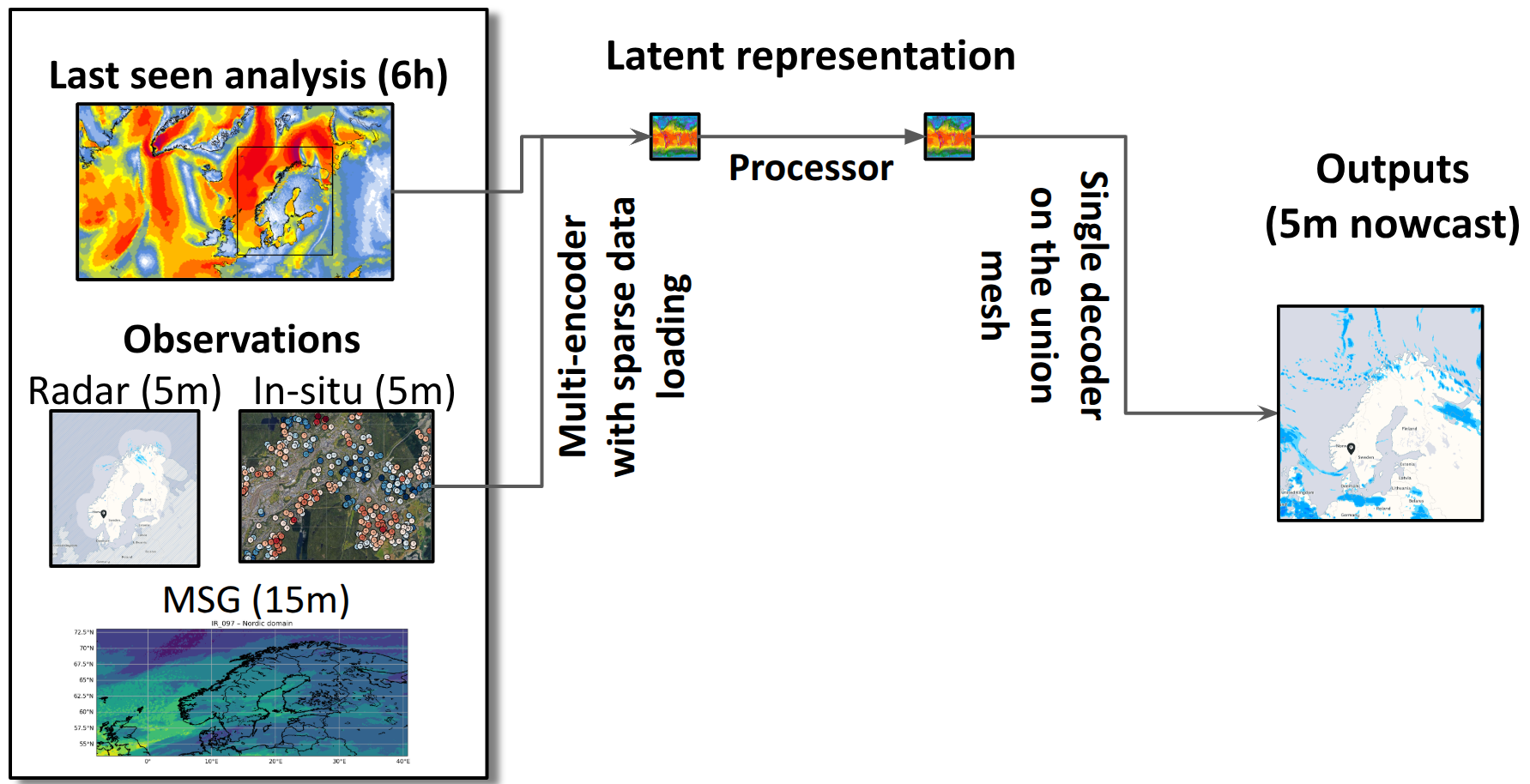}
\caption{Schematic of the multimodal graph neural network nowcaster. Radar, station observations, MSG satellite channels and the most recent NWP state are encoded on their native supports and loaded into a shared latent representation. The processor exchanges information on the graph and a single decoder produces 5~min precipitation nowcasts on the radar grid.}
\label{fig:architecture}
\end{figure}

All models are implemented in the \texttt{anemoi} \citep{anemoi2024} framework and follow the encoder--processor--decoder structure used in recent graph-based weather models. Each data source is represented on its own support: radar and satellite on dense grids, NWP on its native or interpolated grid, and stations on sparse point nodes. Source-specific encoders map these inputs to a shared latent representation (Figure~\ref{fig:architecture}). A graph processor then exchanges information across neighbouring nodes, and a decoder maps the latent state to the radar target grid (for runs without stations truth) or to the union mesh formed by radar grid points and stations locations. The model is trained on a fixed Nordic regional domain. No explicit lateral-boundary treatment is applied. Near the edge of the radar domain, the model relies on the available graph connectivity and on MEPS fields, which provide the large-scale context beyond areas with valid radar coverage. Verification is restricted to valid radar and station locations inside the evaluation mask.

\subsection{Sparse multimodal loading}
\label{sec:sparse}
The implementation is designed to avoid materialising a single dense dataset at the finest available frequency. Radar and Netatmo provide 5min inputs, MSG contributes 15min satellite fields, and MEPS is available at a much coarser 6h frequency. In the configuration used for the multimodal experiments with all data sources, the model encodes 10 recent radar frames, five recent Netatmo frames, four MSG satellite frames, and the most recent available MEPS state and predicts 24 future 5min radar and station steps. Each source's history was chosen to limit the cost of the sparse encoder while encoding as much recent information as possible. The graph contains source-specific nodes for MEPS, radar, Netatmo and MSG, a 4km resolution hidden regional mesh, and a union decoder mesh for radar and station outputs.

This sparse loading is important in practice to achieve better training performance. If all sources are first expanded to the radar frequency and stored as a single physical Zarr array, most of the loaded values are either redundant or unused and training speed becomes dominated by data access. These experiments require dataset-specific input and target offsets for multi-source, multi-frequency training, implemented in \texttt{anemoi-core} as part of this work. This functionality allows each modality to be read only at the time indices required by the corresponding encoder and loss term, rather than forcing all sources onto the highest-frequency timeline. The ablation study would have been too long to run without this feature: radar, stations, satellite fields, and NWP states can be combined without requiring dense 5-min loading of all predictors and targets. Sparse loading increased throughput from approximately 0.03 to 0.20 iterations per second for e.g. the ``+ MEPS" model, corresponding to a factor of about 6--7 in training speed. Equivalently, the time per iteration decreased by roughly 85\% compared with loading all sources after interpolation to the highest temporal frequency.

\subsection{Loss functions}\label{sec:loss}
The loss used for the stochastic multimodal configuration is a weighted combination of four terms on the Nordic radar output, plus an ensemble score on the Netatmo precipitation target. The radar loss is
\begin{equation}
  \mathcal{L}_{\mathrm{rad}} =
  w_1\,\mathcal{L}_{\mathrm{AFKCRPS}}
  +w_2\,\mathcal{L}_{\mathrm{wet}}
  +w_3\,\mathcal{L}_{\mathrm{roll}}
  +w_4\,\mathcal{L}_{\mathrm{SpectralFFT2D}},
  \label{eq:radar_loss}
\end{equation}
where all components are scaled with the number of node weights. The loss weights ($w_1=12$, $w_2=23$, $w_3=161$, $w_4=114$ in Equation~\eqref{eq:radar_loss} are chosen to balance the relative contribution of the different terms. When station targets are available, the five loss components are scaled so that each contributes approximately 20\% of the total loss. When only the four radar-based terms are present, the same procedure targets a contribution of approximately 25\% per term. The weights were calibrated from preliminary runs by monitoring the logged relative contribution of each component to the global loss. When one component dominated the total loss or became negligible, the corresponding weight was adjusted in subsequent runs. 

The first and last terms are the almost-fair kernel CRPS and its spectral counterpart, both with kernel exponent $\alpha=1$. The first is applied directly to the rain-rate field, while the second is computed after applying the two-dimensional Fourier transform. Together, they encourage the ensemble to represent a sharp distribution around the verifying radar field while also constraining the spatial scales of the predicted precipitation structures \citep{lang2024aifscrps, nordhagen2025}. The same almost-fair kernel CRPS is used on Netatmo rain rate, with its own station-node and variable scalers. In the deterministic runs, these probabilistic losses are replaced by a Huber loss (see, e.g., \citet{miralles2025deeplearning}). In the non-ensemble configuration, the spectral loss is replaced with a perceptual loss, the structural similarity index (SSIM).
For two image patches $x$ and $y$, the SSIM is
\begin{equation}
\mathrm{SSIM}(x,y)
=
\frac{(2\mu_x\mu_y + C_1)(2\sigma_{xy}+C_2)}
{(\mu_x^2+\mu_y^2+C_1)(\sigma_x^2+\sigma_y^2+C_2)},
\end{equation}
where $\mu_x$ and $\mu_y$ are local means, $\sigma_x^2$ and $\sigma_y^2$ are local variances, $\sigma_{xy}$ is the local covariance between $x$ and $y$, and $C_1,C_2>0$ are small constants used for numerical stability. SSIM was originally introduced as a perceptual image-quality metric. Rather than comparing images only pixel by pixel, it measures local similarity in luminance, contrast, and structure, thereby approximating aspects of the visual similarity perceived by a human observer. For precipitation nowcasting, we found that log-spectral distance and SSIM lead to broadly similar optima. Both losses penalize errors in spatial structure rather than only pointwise amplitude: SSIM does so through local contrasts and structural similarity in physical space, whereas the log-spectral distance does so by comparing the distribution of power across spatial scales in Fourier space. In practice, both terms encourage coherent precipitation objects and reduce overly noisy or spatially inconsistent fields, although they act through different representations of the same underlying spatial information.

The wet-area term is a differentiable dry/wet mask loss. For a threshold $u$ and temperature $T$, define
\begin{equation}
q_u(x) = \sigma\left(\frac{x-u}{T}\right),
\end{equation}
where $\sigma$ is the logistic function. For each ensemble member $m$, the loss compares the soft wet mask of the prediction with the observed wet mask and can weight false positives and false negatives separately,
\begin{equation}
\mathcal{L}_{\mathrm{wet}}^{(m)} =
w_+\left\{q_u(\hat y^{(m)})-q_u(y)\right\}_+^2
+
w_-\left\{q_u(y)-q_u(\hat y^{(m)})\right\}_+^2 .
\end{equation}
The member-wise losses are then aggregated over the ensemble. In the configuration used here $u=0.1$~mm~h$^{-1}$ and $T=0.03$. This term is included because small errors in the rainfall rate around the dry/wet boundary can produce visually and operationally important false rain areas. 

The rolling accumulation term $\mathcal{L}_{\mathrm{roll}}$ is a Huber loss on short accumulations, also computed separately for each ensemble member before aggregation. With output interval $\Delta t=300/3600$~h and window length $W=3$, the predicted accumulation for member $m$ is
\begin{equation}
A_{t}^{(m)} = \sum_{j=0}^{W-1} \Delta t,\hat y_{t+j}^{(m)} .
\end{equation}
The target accumulation is defined in the same way, and $\mathcal{L}_{\mathrm{roll}}$ applies a Huber loss with $\delta=0.3$ to the difference between predicted and observed rolling accumulations. This term stabilizes the 15min precipitation amount, and is useful because a nowcast that is noisy at each 5min step can still be wrong for the short accumulations used by forecasters.

To separate the effect of stochastic regularisation from that of probabilistic training, the ``+ noise'' run uses the same noise input and the same number of members as the CRPS run, but replaces the ensemble score with a deterministic loss applied member-wise. This control experiment clarifies the source of any improvement. If the noise-only model performs similarly to the CRPS model, the gain is likely due mainly to noise augmentation or ensemble aggregation. If, however, the CRPS model further improves calibration, spread--error consistency, high-intensity precipitation, or probabilistic scores, then the ensemble objective is contributing beyond the regularising effect of noise.

\subsection{Ablation design}
\label{sec:ablation}
\begin{table}[tbp]
\centering
\small
\setlength{\tabcolsep}{3.5pt}
\renewcommand{\arraystretch}{1.15}
\caption{Summary of the ablation runs. Predictors denote data sources available to the model at initialization time. Targets denote the future data sources used for supervision during training. The run names are cumulative: each row adds one modelling component or training objective to the previous deterministic configuration. Runtimes are measured for a two-hour forecast. For ensemble runs, runtimes are reported per member and, when members are run serially, for the full ensemble.}
\label{tab:runs}
\begin{tabularx}{\textwidth}{
  >{\raggedright\arraybackslash}p{0.20\textwidth}
  >{\raggedright\arraybackslash}p{0.24\textwidth}
  >{\raggedright\arraybackslash}p{0.14\textwidth}
  >{\centering\arraybackslash}p{0.06\textwidth}
  >{\raggedright\arraybackslash}p{0.16\textwidth}
  >{\raggedright\arraybackslash}X
}
\toprule
Run & Predictors & Targets & Ens. & Model calls & Runtime \\
\midrule
Optical flow benchmark
  & \texttt{radar}
  & \texttt{radar}
  & 1
  & 1
  & 15.0 s \\
\hline 
Radar DL baseline
  & \texttt{radar}
  & \texttt{radar}
  & 1
  & 24; rollout every 5 min
  & 60.0 s \\
\hline
+ MEPS
  & \makecell[l]{\texttt{radar}\\\texttt{MEPS}}
  & \texttt{radar}
  & 1
  & 4; rollout every 30 min
  & 8.5s \\
\hline
+ MEPS + Netatmo
  & \makecell[l]{\texttt{radar}\\\texttt{MEPS}\\\texttt{Netatmo}}
  & \makecell[l]{\texttt{radar}\\\texttt{Netatmo}}
  & 1
  & 4; rollout every 30 min
  & 8.3s \\
\hline
+ MEPS + Netatmo + satellite
  & \makecell[l]{\texttt{radar}\\\texttt{MEPS}\\\texttt{Netatmo}\\\texttt{MSG}}
  & \makecell[l]{\texttt{radar}\\\texttt{Netatmo}}
  & 1
  & 4; rollout every 30 min
  & 10.6s \\
\hline
+ MEPS + Netatmo + satellite + noise
  & \makecell[l]{\texttt{radar}\\\texttt{MEPS}\\\texttt{Netatmo}\\\texttt{MSG}\\\texttt{noise}}
  & \makecell[l]{\texttt{radar}\\\texttt{Netatmo}}
  & 15
  & 4; rollout every 30 min
  & 10.45 s/member; 156.75s serial \\
\hline
+ MEPS + Netatmo + CRPS
  & \makecell[l]{\texttt{radar}\\\texttt{MEPS}\\\texttt{Netatmo}\\\texttt{noise}}
  & \makecell[l]{\texttt{radar}\\\texttt{Netatmo}}
  & 15
  & 4 per member; rollout every 30 min
  & 9.96 s/member; 149.4s serial \\
\hline
+ MEPS + Netatmo + satellite + CRPS
  & \makecell[l]{\texttt{radar}\\\texttt{MEPS}\\\texttt{Netatmo}\\\texttt{MSG}\\\texttt{noise}}
  & \makecell[l]{\texttt{radar}\\\texttt{Netatmo}}
  & 15
  & 4 per member; rollout every 30 min
  & 161.4 s serial \\
\bottomrule
\end{tabularx}
\end{table}
The external baseline model is a deterministic optical-flow nowcasting model based on the advection of recent radar fields \citep{ipol201326}. Deep learning models are compared against one another in an ablation setting summarised in Table~\ref{tab:runs}. The training period spans over 4 years (May 2020 to August 2024); validation is only run on the few remaining months from 2024 and 2025 is used as the test set.
The number of model calls differs between runs (Table~\ref{tab:runs}) because the radar-only model was more sensitive to the parallel output structure. The network predicts six 5-min precipitation fields per forward pass. When the radar-only model was rolled out only every six output steps, the resulting sequence showed visible discontinuities every 30min, corresponding to the transition between consecutive forecast blocks. To obtain a stronger and fairer radar-only benchmark, we therefore rolled this model out every 5-min step, leading to 24 model calls for a two-hour forecast. Adding external predictors alleviated this constraint. With NWP, station, and satellite information, the six parallel output steps within each forecast block were less similar to each other, precipitation structures moved more smoothly, and the 30-min block transitions no longer produced the same visible jumps. The multimodal runs were therefore rolled out every 30 min, reducing inference cost without introducing temporal artefacts. The target variable is instantaneous rain rate, denoted by $y_t(s)$ at time $t$ and grid point $s$. The model predicts a sequence of future rain-rate fields every five minutes for the two next hours. Radar quality-control flags (see Section~\ref{sec:data}) are used to mask missing or unreliable grid points during training and verification.

\section{Verification framework}\label{sec:verification}
The verification is designed to separate several failure modes that can otherwise be conflated when only pointwise scores are used. We first compute standard pointwise metrics, including RMSE, MAE, bias, Pearson correlation, and wet-point RMSE, both against the radar grid and against the nearest Netatmo stations. The radar-grid scores evaluate whether the model predicts the spatial field it was trained to produce, whereas the station scores evaluate whether the predicted field is locally consistent with independent point observations. 

Spatial structure is further evaluated with SSIM applied to $\log(1+y)$ rain-rate fields. The logarithmic transform reduces the dominance of the largest precipitation intensities and makes the score more sensitive to the structure of weak and moderate rain. SSIM complements pixelwise scores because it compares local means, contrasts, and covariance rather than independent grid-point errors. It therefore provides a diagnostic of precipitation morphology: high log-rain SSIM indicates that the forecast preserves the local organisation and texture of the radar field, even when some pointwise amplitude errors remain.

We also evaluate rain-onset skill at Netatmo stations, because short-range nowcasting is often used to answer whether rain will start soon at a given location. Onset cases are defined only for stations that are dry at the forecast reference time. For each such station, we examine the following two-hour forecast window. An observed onset event occurs if the total observed rain over this window exceeds 0.1mm. The observed onset time is then the first 5min time step at which rain is detected at the station. The predicted onset time is defined analogously from the forecast evaluated at the nearest prediction grid point. For stations with an observed onset and a predicted onset, we compute the onset bias and RMSE in minutes. A miss occurs when rain is observed during the two-hour window but the forecast remains dry throughout the window. A false alarm occurs when the station remains dry over the full two-hour window but the forecast predicts rain at least once. The miss rate therefore measures failure to predict rain initiation, whereas the false-alarm rate measures whether the model creates spurious local rain signals.

Finally, we use displacement and amplitude diagnostics to determine whether remaining errors are primarily caused by misplaced precipitation, mistimed precipitation, or wrong intensities. The spatial-oracle diagnostic searches over a set of domain-wide translations of the forecast field and reports the best score after shifting the prediction. The relative improvement obtained by this oracle shift measures how much of the error could be removed by a simple spatial correction. The temporal-oracle diagnostic applies the same idea in time, by comparing the forecast to radar fields at neighbouring valid times and measuring the improvement obtained by a short temporal lag. Large oracle gains would indicate that the model predicts realistic precipitation structures but places them at the wrong position or time. Small oracle gains, in contrast, suggest that the main limitation lies in local structure or amplitude rather than in bulk displacement.

The DAS-lite diagnostic provides a more explicit decomposition of this effect. It uses the best domain-wide translation from the spatial-oracle search as a computationally cheap approximation to the displacement component of the displacement and amplitude score. The displacement component is the magnitude of the best shift, reported in kilometres, while the amplitude component is the wet-area RMSE after applying this shift, normalized by a characteristic observed intensity. This is not a full dense optical-flow DAS \citep{keil2009das}, but it is useful as a diagnostic: a high displacement component points to advection or motion errors, and a high amplitude component after alignment might indicate errors linked to the intensity or structure of the predicted rain field. 

These diagnostics help identify which failure modes should be targeted in follow-up experiments. If the dominant error is displacement, the next step would be to improve motion modelling or rollout dynamics. If station scores improved but radar-grid SSIM did not, the next step would be to investigate station memorisation and sensor-geometry generalisation. If displacement and timing gains remain small while amplitude errors dominate, as in the current experiments, then the natural follow-up is a generative or guided-assimilation approach that can use stations as constraints on a distribution of radar-like precipitation fields rather than as additional pointwise regression targets.

\section{Results}
\label{sec:results}
The ablation does not show a monotonic improvement in which each additional data source improves all scores. Instead, each component affects a different property of the forecast. We therefore interpret the results separately for radar-grid skill, station and onset behaviour, the effect of direct Netatmo fusion, and the overall balance between spatial accuracy and event realism.

\subsection{Ablation results by forecast property}
\begin{figure}[ht!]
\centering
\includegraphics[width=\textwidth]{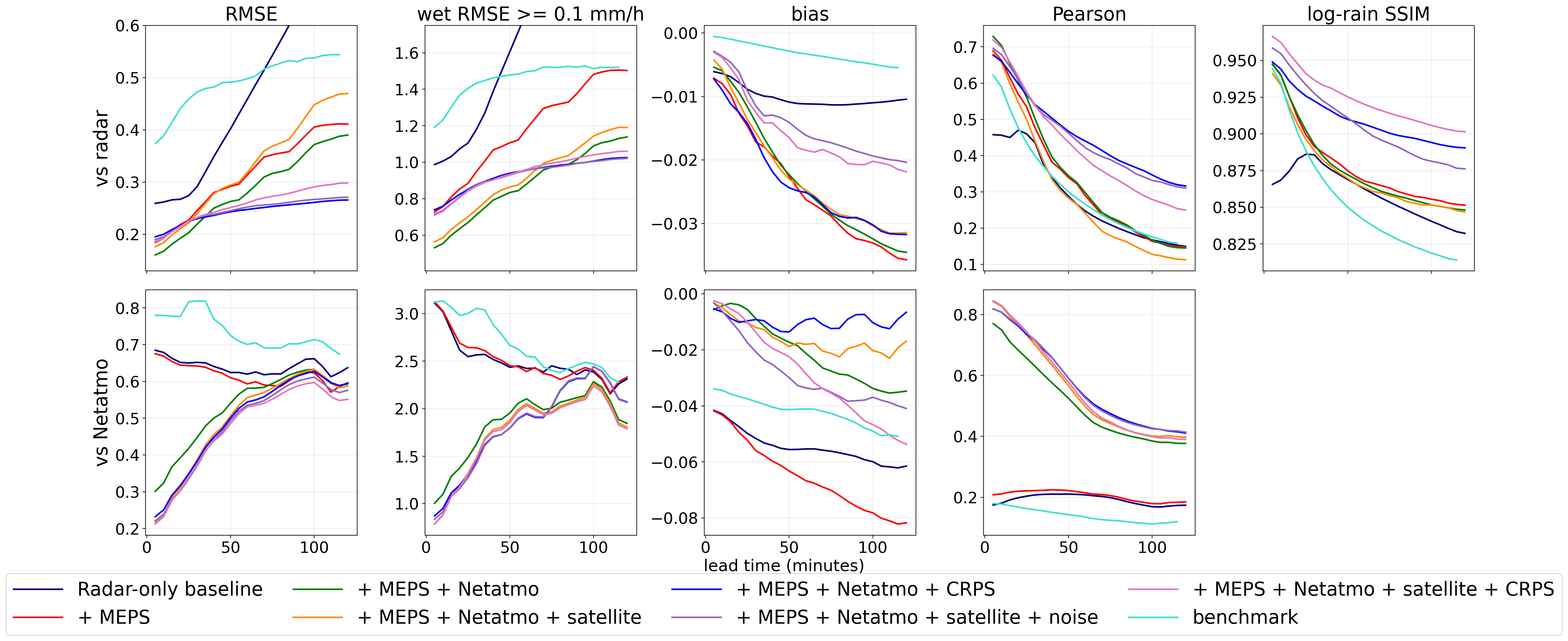}
\caption{Pixelwise and pointwise verification as a function of lead time. The first row is evaluated against the Nordic radar target and the second row against Netatmo observations. The ranking of the models depends strongly on the verification support. CRPS-based training gives the most consistent improvement against the radar grid, whereas station-based verification emphasizes local agreement and can favour runs that do not clearly improve the dense precipitation field.}
\label{fig:metrics}
\end{figure}

\paragraph{Radar-grid field skill}
The most consistent improvement against the radar target comes from the CRPS-based runs (Figure~\ref{fig:metrics}). The + CRPS configurations both with and without satellite inputs reduce radar-grid RMSE and wet RMSE relative to the radar-only and MEPS-only models, and maintain higher correlation at longer lead times. This suggests that the probabilistic objective contributes beyond simply adding stochastic noise or additional predictors. The radar-only model provides a useful baseline, but its error grows rapidly with lead time. MEPS stabilizes the forecast to some extent, but does not by itself produce the same level of improvement as the CRPS-based configurations. The optical-flow benchmark remains competitive at the shortest lead times, but its larger RMSE and weaker structural scores at later lead times indicate the limitations of pure advection.
\begin{figure}[ht!]
\centering
\includegraphics[width=\textwidth]{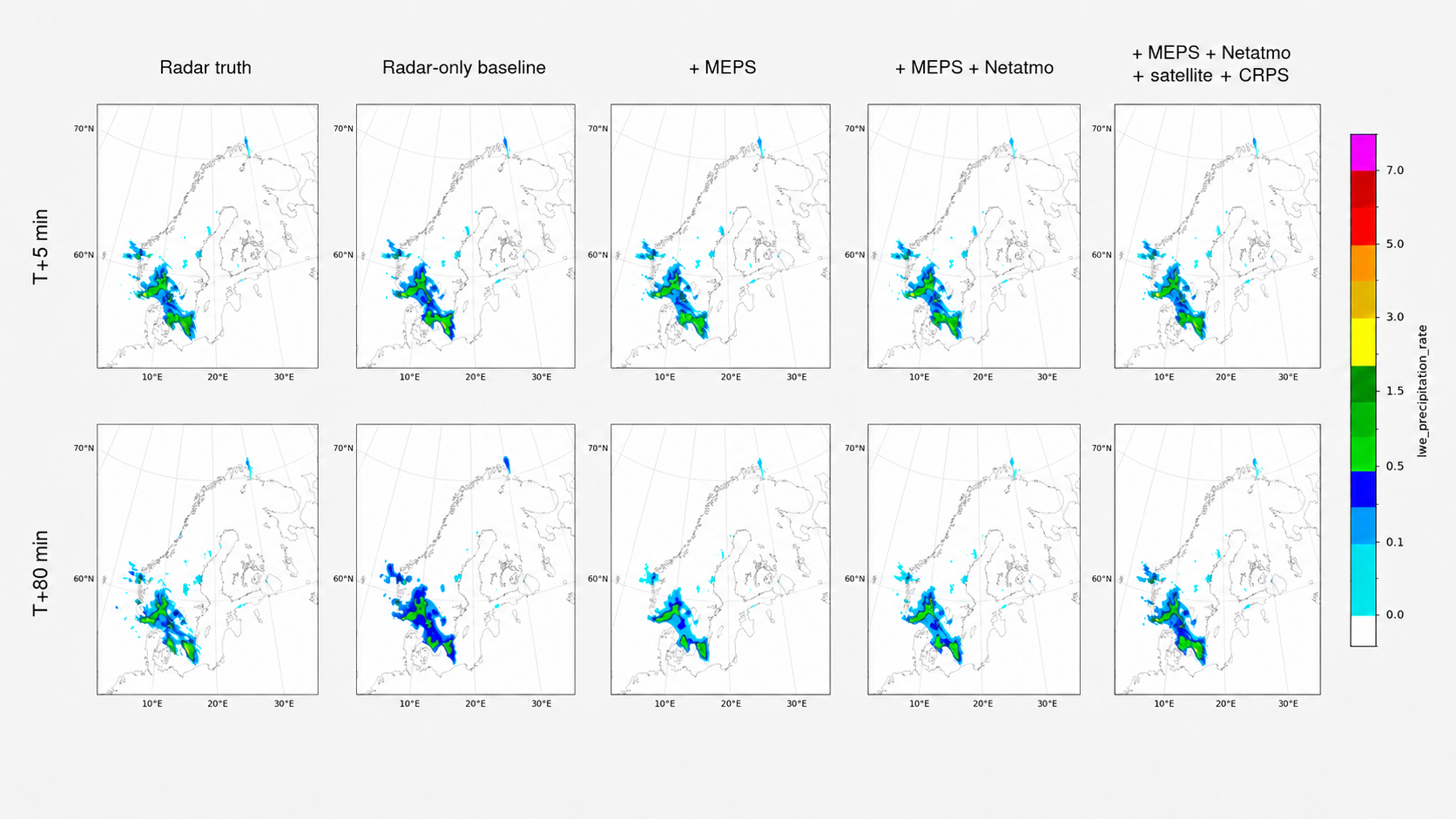}
\caption{Example radar-field evolution for a representative precipitation case at T+5~min and T+80~min. Columns compare the radar truth, the radar-only baseline, the MEPS-informed run, the MEPS+Netatmo run, and the MEPS+Netatmo+satellite+CRPS run. The figure illustrates that all learned configurations capture the main precipitation area at short lead time, while differences in spatial organisation and amplitude become clearer at longer lead time.}
\label{fig:case_radar_field}
\end{figure}
Figure~\ref{fig:case_radar_field} illustrates the evolution of a representative case at T+5~min and T+80~min. At the first lead time, all learned configurations reproduce the main precipitation structure, but the radar-only baseline already shows a smoother and more spatially consolidated field than the radar truth. At T+80~min, differences between configurations become clearer. The MEPS-informed and station-informed runs retain the broad location of the precipitation system, while the satellite+CRPS configuration gives a more coherent precipitation object and a spatial structure closer to the radar field. This example is consistent with the aggregate radar-grid diagnostics: most configurations capture the large-scale rain area, but the probabilistic satellite-informed run better preserves the spatial organisation and amplitude of the precipitation field at longer lead times.

\begin{figure}[ht!]
\centering
\includegraphics[width=\textwidth]{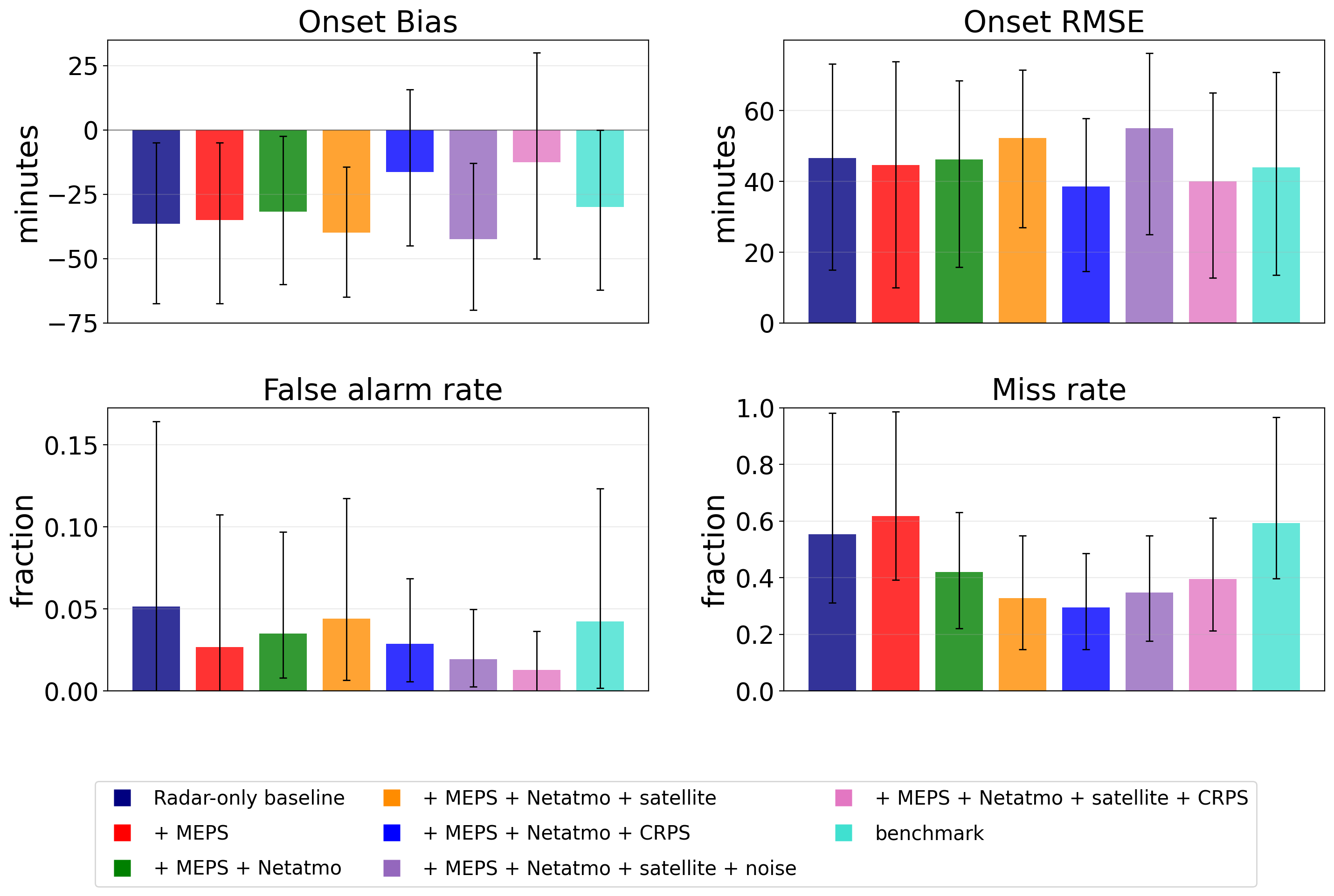}
\caption{Rain-onset diagnostics against Netatmo observations. Deterministic satellite-guided forecasts reduce some misses but also trigger precipitation too early and too often, as shown by the negative onset bias and high false-alarm rate. The CRPS-based configurations generally provide a more balanced treatment of rain initiation, with lower radar-grid errors and less severe onset miscalibration.}
\label{fig:onset_metrics}
\end{figure}

\begin{figure}[p]
\centering
\includegraphics[width=0.95\textwidth]{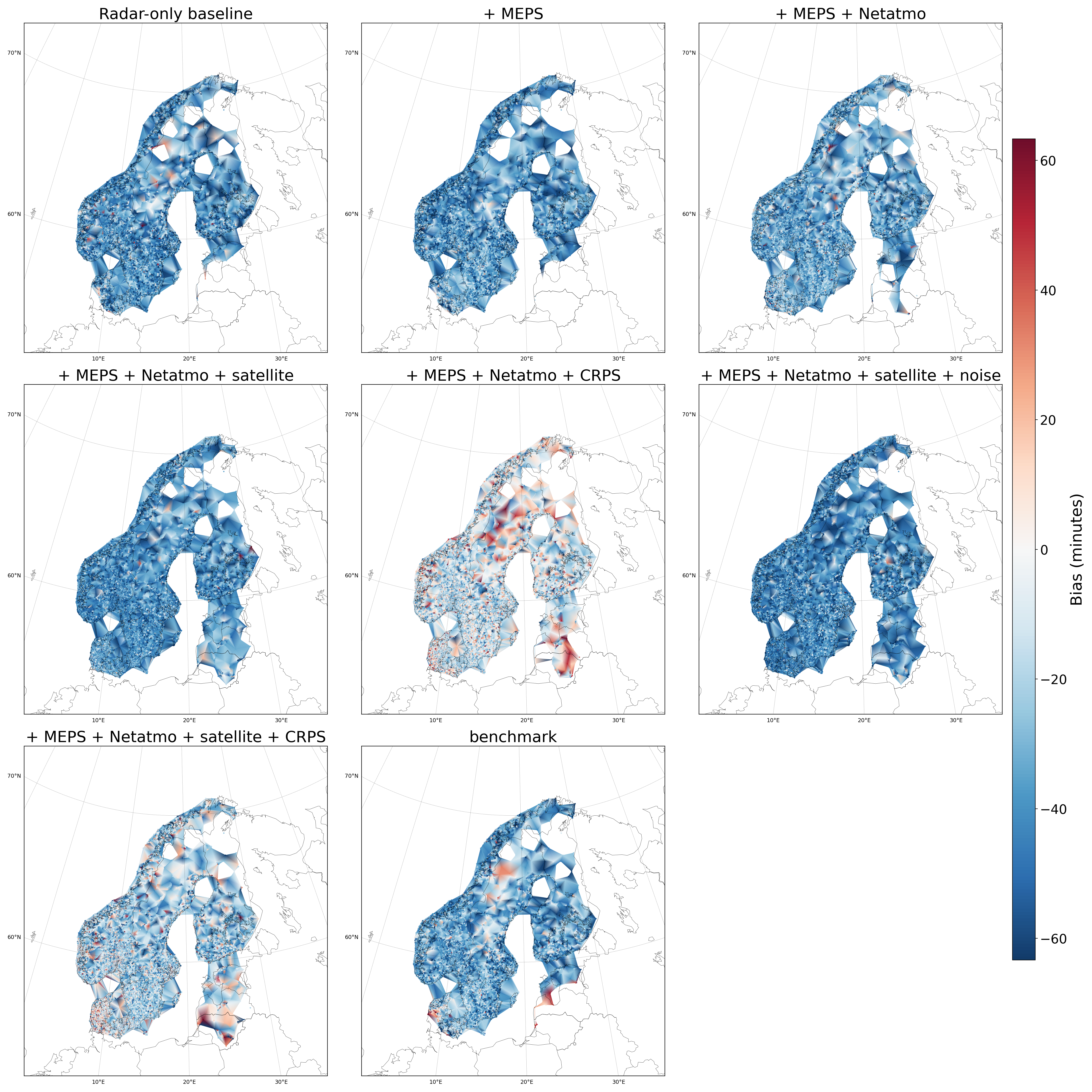}
\caption{Spatial distribution of rain-onset bias at Netatmo stations. Negative values indicate that the forecast predicts rain too early, while positive values indicate delayed onset. The runs using satellite data as a predictor shows a broad early-onset tendency, whereas the CRPS-based configurations reduce the domain-wide early bias and produce a more balanced spatial pattern.}
\label{fig:onset_bias_map}
\end{figure}

\begin{figure}[p]
\centering
\includegraphics[width=0.95\textwidth]{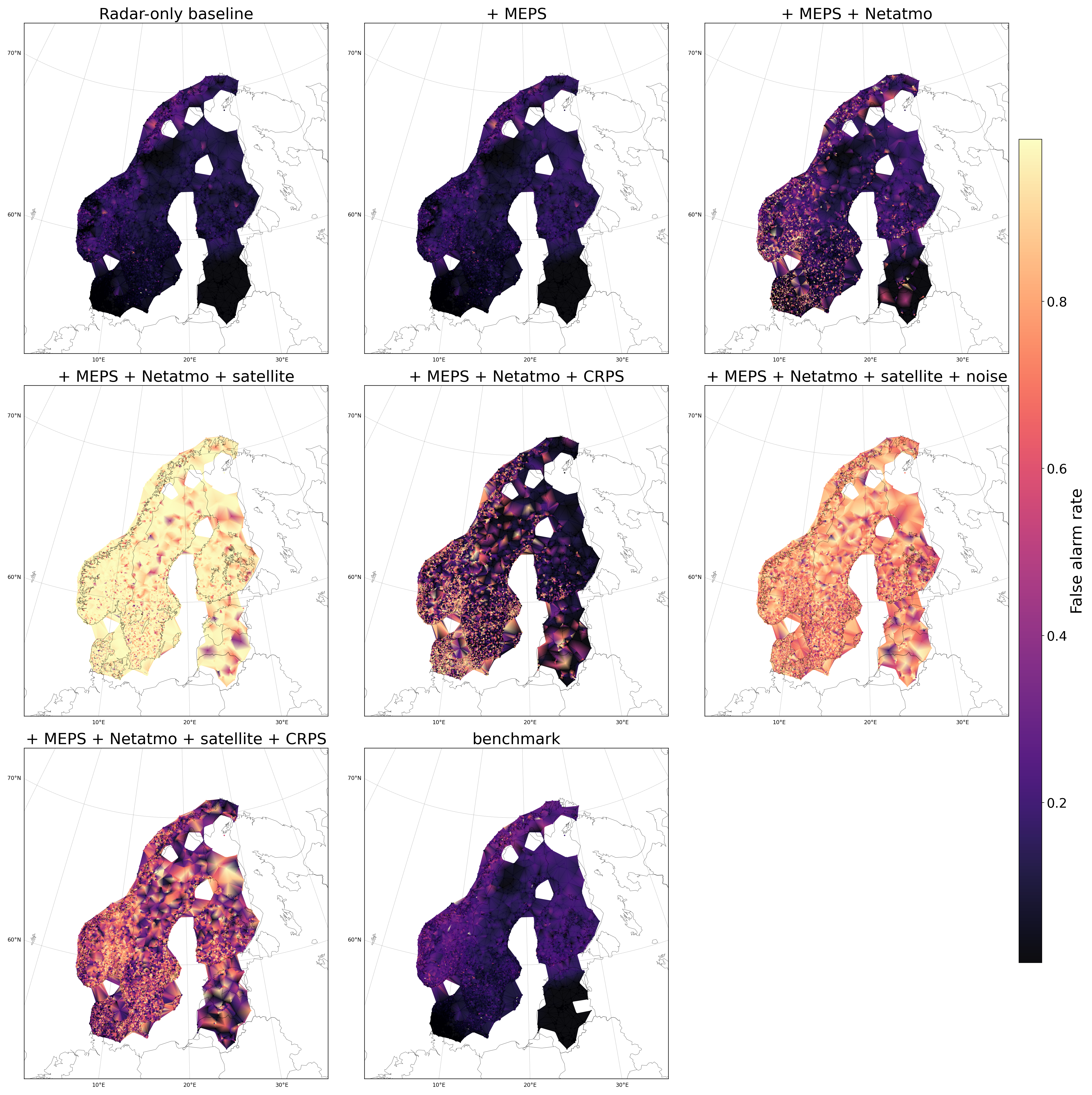}
\caption{Spatial distribution of rain-onset false-alarm rate at Netatmo stations. False alarms occur when the station remains dry over the two-hour window but the forecast predicts rain at least once. The deterministic satellite configuration produces widespread false alarms, suggesting that satellite predictors help identify precipitation-prone cloud structures but can over-activate rain in a deterministic setting.}
\label{fig:onset_far_map}
\end{figure}

\paragraph{Onset behaviour}
All configurations tend to predict precipitation too early on average (Figure~\ref{fig:onset_metrics}). The stochastic run has the smallest onset bias (Figure~\ref{fig:onset_bias_map}) and the lowest false alarm rate (Figure~\ref{fig:onset_far_map}), but its uncertainty is large.
The spatial distribution of onset bias is shown in Figure~\ref{fig:onset_bias_map}. The deterministic runs are dominated by negative biases over most of the station network. MEPS reduces some of the broad spatial structure of the bias, while the stochastic run is less uniformly early but introduces more local positive and negative patches. 

\begin{figure}[p]
\centering
\includegraphics[width=0.95\textwidth]{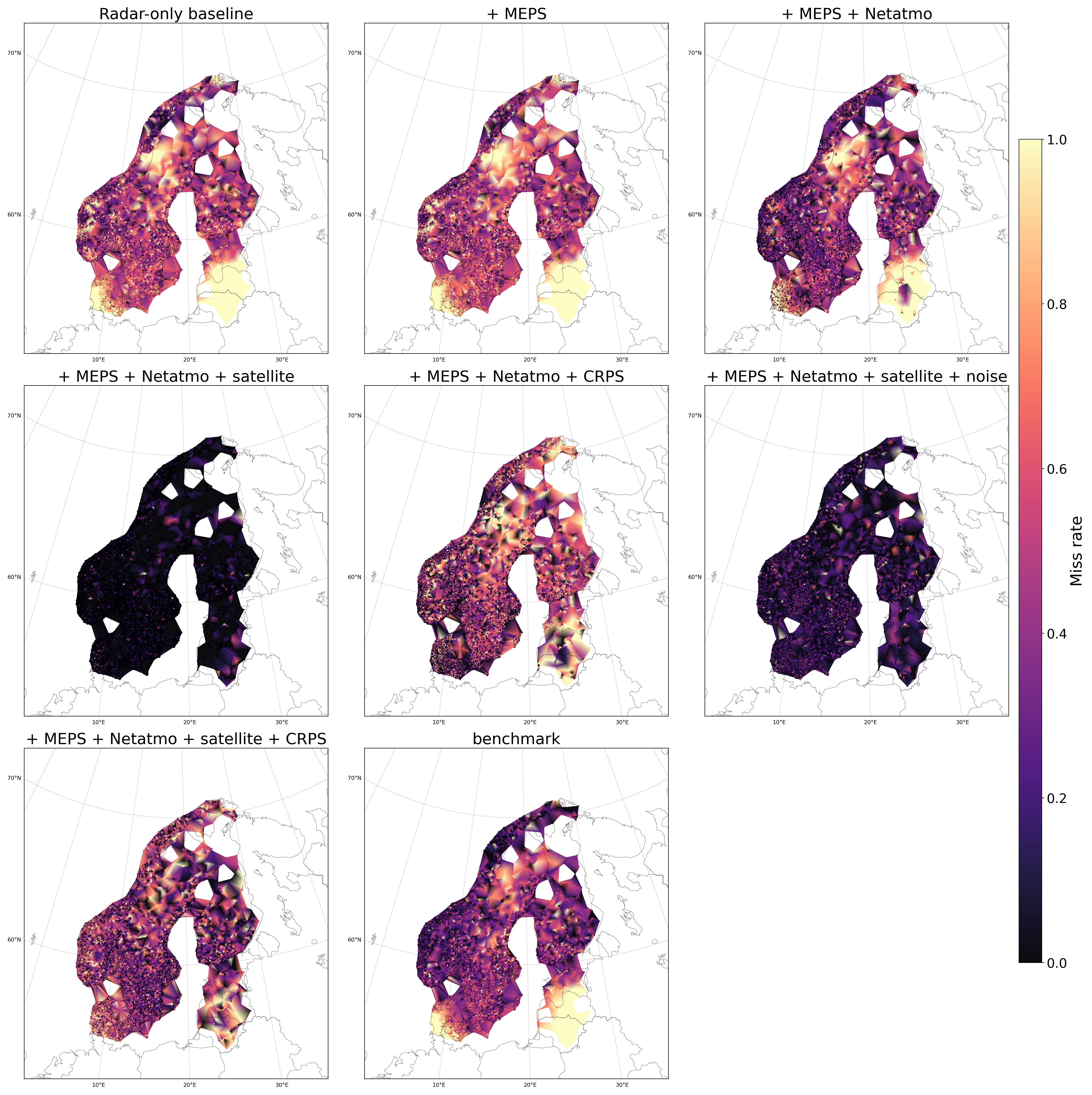}
\caption{Spatial distribution of rain-onset miss rate at Netatmo stations. Misses occur when rain is observed during the two-hour window but the forecast remains dry. The miss maps complement the false-alarm maps by showing whether a model is conservative or over-active in predicting new rain events.}
\label{fig:onset_miss_map}
\end{figure}

\paragraph{Satellite predictors}
Satellite information contains useful spatial context about cloud and precipitation systems, and this is visible in the reduced bias against Netatmo observations for the satellite-informed runs. (Figure~\ref{fig:metrics}). However, this improved station-level rain-rate bias does not imply better calibration of rain initiation. Both runs using satellite with or without input noise tend to trigger rain too early and too often, with a strongly negative onset bias and a high false-alarm rate (Figure~\ref{fig:onset_metrics}). 

The spatial onset maps provide a more detailed view of the tradeoff between missed rain initiation and false alarms. The deterministic satellite run has a clear early-onset signature, visible as widespread negative onset bias and elevated false-alarm rates (Figures~\ref{fig:onset_bias_map} and~\ref{fig:onset_far_map}). This supports the interpretation that satellite channels provide useful information about cloud and precipitation organisation, but that in a deterministic model this information can be converted into rain too aggressively. The CRPS-based configurations reduce this imbalance. They do not remove all local timing errors, but they avoid the broad false-alarm behaviour of the deterministic satellite run while retaining improved radar-grid skill.
The miss-rate maps show the complementary side of the same tradeoff (Figure~\ref{fig:onset_miss_map}): A model can reduce missed onsets by predicting rain more often, but this is only useful if it does not also increase false alarms.

\paragraph{Direct Netatmo fusion}
The station-based diagnostics indicate that the Netatmo observations provide useful local information in the present setup (Figure~\ref{fig:metrics}). Runs that include Netatmo tend to improve the verification against station observations, including pointwise rain-rate metrics and onset diagnostics. This suggests that the model is able use the additional surface observations at or near the locations where they directly constrain the forecast. In particular, the station-informed runs appear to better represent local rain/no-rain occurrence and precipitation reaching the surface, which are precisely the aspects for which point observations are expected to be informative (Figure~\ref{fig:onset_metrics}).

At the same time, the improvement at station locations does not translate systematically into improved radar-grid scores (Figure~\ref{fig:metrics}). This distinction is important because the two verification supports measure different properties of the forecast. Station verification evaluates local agreement with surface precipitation observations, whereas radar-grid verification evaluates the spatial organisation of a dense radar-like field. The results are therefore consistent with a scale-support mismatch: Netatmo observations add valuable local constraints, but those constraints do not uniquely determine how the neighbouring radar field should be modified.

\paragraph{Best overall behaviour}
No configuration dominates all diagnostics. The + CRPS only run is the cleanest evidence that the ensemble objective improves the forecast, because it performs well without relying on satellite predictors. The + CRPS + satellite run gives the most balanced behaviour: it retains much of the radar-grid improvement from CRPS-based training while using satellite information without the severe false-alarm behaviour of the deterministic satellite run. 
Overall, the ablation shows that multimodal learning improves precipitation nowcasting in many different aspects. MEPS provides dynamical context, satellite channels provide information on cloud and precipitation organisation, and CRPS-based training improves the treatment of uncertainty and precipitation amplitudes. Direct station fusion remains the least resolved part of the system: it improves local pointwise agreement without consistently improving the surrounding radar field. 

\subsection{Oracle, displacement, and amplitude diagnostics}
The oracle and DAS-lite diagnostics suggest that the main remaining errors are not large-scale displacement errors. Spatial oracle gains are below 1\% for all runs, indicating that simple domain-wide shifts cannot substantially improve the forecasts. Temporal oracle gains are larger, but remain moderate, with the best temporal correction corresponding to a short negative lag. The main discriminating diagnostic is the amplitude component after alignment. Adding MEPS and Netatmo reduces the amplitude error relative to the radar-only model, while the CRPS-only and satellite+noise configurations give much lower amplitude components. The deterministic satellite run does not improve this diagnostic, suggesting that satellite predictors alone are not sufficient; their benefit appears mainly when combined with either stochastic perturbations or the CRPS-based ensemble objective. The combined satellite+CRPS configuration gives the lowest overall score which can be interpreted as satellite information and probabilistic training contributing to complementary improvements.

\begin{table}[ht!]
\centering
\caption{Oracle and DAS-lite diagnostics. Lower values are better for the overall score, DAS displacement, and DAS amplitude. Large oracle gains would indicate that displacement or timing errors dominate the forecast error. Spatial and temporal oracle gains are reported as relative improvements over the original score.}
\label{tab:diagnostics}
\resizebox{\textwidth}{!}{%
\begin{tabular}{lccccc}
\toprule
Run & Overall score & Spatial oracle gain & Temporal oracle gain & DAS displacement & DAS amplitude \\
 & $\downarrow$ & (\%) $\uparrow$ & (\%) $\uparrow$ & km $\downarrow$ & $\downarrow$ \\
\midrule
Radar only
  & 0.157
  & 0.58
  & 4.29
  & 2.57
  & 0.764 \\
+ MEPS
  & 0.168
  & 0.03
  & 2.59
  & 0.73
  & 0.513 \\
+ MEPS + Netatmo
  & 0.167
  & 0.07
  & 2.33
  & 1.03
  & 0.497 \\
+ MEPS + Netatmo + satellite
  & 0.178
  & 0.05
  & 2.88
  & 0.93
  & 0.574 \\
+ MEPS + Netatmo + CRPS
  & 0.179
  & 0.03
  & 2.71
  & 1.54
  & 0.336 \\
+ MEPS + Netatmo + satellite + noise
  & 0.162
  & 0.07
  & 2.51
  & 1.76
  & 0.328 \\
+ MEPS + Netatmo + satellite + CRPS
  & 0.148
  & 0.19
  & 3.50
  & 1.15
  & 0.335 \\
\bottomrule
\end{tabular}%
}
\end{table}

\section{Conclusion}\label{sec:discussion}
The ablation indicates that multimodal fusion is not simply a matter of adding more predictors. The best component depends on the forecast property being evaluated: radar-grid error, station agreement, onset timing, false alarms, or amplitude after alignment. MEPS provides spatially organised dynamical information and stabilises the forecast relative to the radar-only model. Satellite predictors provide useful information about cloud and precipitation organisation, and this is visible in some radar-grid diagnostics and in the reduced bias against Netatmo observations for the satellite-informed runs. However, the onset diagnostics show that this information is not automatically beneficial: when used without a probabilistic objective, satellite information can lead to premature or excessive rain activation.

The CRPS-based configurations provide the most consistent improvement against the radar target. The CRPS-only run is particularly useful as a control, because it shows that the ensemble objective contributes even without satellite predictors. The combined satellite+CRPS configuration gives the most balanced behaviour overall: it retains the spatial information provided by satellite channels while avoiding the strongest false-alarm behaviour seen in the satellite-informed deterministic or noise-only configurations. The DAS-lite diagnostics lead to a similar interpretation. The spatial-oracle gains are small for all runs, suggesting that the dominant errors are not simple bulk displacement errors. The strongest differences appear in the amplitude component after alignment, where the stochastic and CRPS-based configurations substantially reduce the error relative to the radar-only, MEPS, and direct Netatmo runs.

Netatmo stations provide direct information about precipitation reaching the ground, but their spatial support is a point. If the training objective treats a station observation as another target value in the same supervised loss, the model can improve the exact station location without learning how the surrounding radar-like field should be modified. This does not mean that Netatmo observations are uninformative. Their added value might not fully realised using direct pointwise fusion. Indeed, a wet or dry station signal constrains the local precipitation state, but many spatially coherent radar fields remain compatible with the same signal.
This result has consequences for the design of precipitation and lightning nowcasting systems. Lightning observations, like stations, are sparse and intermittent. A lightning detection is highly informative about convective activity, but it does not determine a unique rain-rate field. Similarly, a station wet/dry signal is informative about surface precipitation at one location, but it does not prescribe the spatial structure of the surrounding precipitation object. These observations are therefore better interpreted as constraints on plausible precipitation states than as ordinary regression targets.

The present GNN implementation also becomes heavier as more data sources and ensemble members are added. This is important for operational use. The obvious next modalities over the Nordics include polar-orbiting satellite instruments, lightning networks, additional radar volumes, radar quality indicators, and uncertainty estimates. Future architectures may therefore need to combine efficient image-based encoders for dense sources with sparse graph nodes for stations and lightning. In that setting, the key methodological question is not only how to ingest more observations, but how to use observations with different spatial supports to update a coherent probabilistic precipitation field.

Overall, the experiments motivate a follow-up study on guided generative assimilation. In such a framework, stations and lightning observations would be treated as likelihood constraints on a distribution of plausible precipitation analyses, rather than as additional pointwise targets competing with the radar field. This would allow sparse observations to modify radar-like precipitation states probabilistically while preserving spatial coherence.

\section*{Acknowledgments}
Computational resources, data access, and operational context were provided by the Norwegian Meteorological Institute and EuroHPC through project~465002688. The author thanks colleagues working on Anemoi, Bris, radar processing, station quality control, and operational nowcasting for discussions that motivated this study. 

\newpage
\appendix
\section{Radar Quality Flag Pseudo-Code}\label{app:qc_flags}
\small
\begin{verbatim}
Input:
    rain_rate                  # radar rain-rate field
    native_qc_flags             # Boolean radar QC fields
    block_percent               # beam-blocking percentage

Initialise:
    qc_flags = zeros_like(rain_rate, dtype=uint16)

Encode native radar flags:
    bit 0  <- is_nodata
    bit 1  <- is_blocked
    bit 2  <- is_lowele
    bit 3  <- is_highele
    bit 4  <- is_seaclutter
    bit 5  <- is_groundclutter
    bit 6  <- is_otherclutter
    bit 7  <- is_convective

for each flag index k:
    qc_flags = qc_flags OR (native_qc_flags[k] << k)

Derive additional flags:
    block_gt50      = is_blocked AND block_percent > 50
    extreme_20_50   = 20 < rain_rate <= 50
    invalid_high    = rain_rate > 50

Encode derived flags:
    bit 8  <- block_gt50
    bit 9  <- extreme_20_50
    bit 10 <- invalid_high

Define hard quality-control mask:
    hard_nan =
        is_nodata OR
        is_seaclutter OR
        is_groundclutter OR
        is_otherclutter OR
        block_gt50 OR
        invalid_high

Apply hard mask:
    rain_rate[hard_nan] = NaN

Store:
    rain_rate
    qc_flags

Decode when needed:
    decoded_flag[k] = (qc_flags >> k) AND 1
\end{verbatim}

\newpage
\bibliographystyle{plainnat}
\bibliography{bibli}

\end{document}